\documentclass{article}

\usepackage[utf8]{inputenc}
\usepackage[T1]{fontenc}
\usepackage{hyperref}
\usepackage{url}
\usepackage{booktabs}
\usepackage{graphicx}
\usepackage{amsmath,amssymb}
\usepackage{microtype}
\usepackage{xcolor}
\usepackage{caption}
\usepackage{subcaption}
\usepackage{svg}
\usepackage[margin=1.2in]{geometry}
\usepackage{times}

\begin{document}

\title{
Toward Lightweight and Fast Decoders for \\
Diffusion Models \\
in Image and Video Generation
}

\author{
    Alexey Buzovkin \\
    \texttt{alexey.buzovkin@gmail.com}\\[6pt]
    \ Evgeny Shilov \\
    \texttt{jvshilov@gmail.com}
}
\date{}

\maketitle

\begin{abstract}
\noindent
We investigate methods to reduce inference time and memory footprint in stable diffusion models by introducing lightweight decoders for both image and video synthesis. Traditional latent diffusion pipelines rely on relatively large Variational Autoencoder (VAE) decoders that can significantly slow down generation and consume considerable GPU memory. We propose custom-trained decoders that utilize lightweight Vision Transformer and Taming Transformer architectures. Our experiments demonstrate up to 15\% overall speed-ups for image generation on COCO2017 (and up to 20$\times$ faster just in the decoding sub-module), with further gains on the UCF-101 dataset for video. Memory requirements are also moderately reduced. Although there is a small drop in perceptual quality compared to the default decoder, the improvement in speed and scalability can be critical for large-scale inference scenarios (e.g., generating 100K images). Our work is further contextualized by advances in efficient video generation, such as VideoMAE V2's dual masking strategy, illustrating a broader effort in the field to improve the scalability and efficiency of generative models.

Examples of inference, metrics evaluation, and further useful materials are located in \url{https://github.com/RedShift51/fast-latent-decoders}.
\end{abstract}

\section{Introduction}
Recent advances in diffusion models have enabled high-fidelity image and video generation in various domains. Models such as Stable Diffusion~\cite{sdxl, sddm} employ a latent diffusion pipeline with two main steps:
\begin{enumerate}
    \item Denoising in latent space (via a UNet backbone).
    \item Decoding from the latent representation to pixel space with a Variational Autoencoder (VAE).
\end{enumerate}
Although these approaches produce impressive visual quality, the decoder can become a bottleneck for real-time or high-throughput applications because of its size and high computational demands. The generation of hundreds of thousands of images or frames further amplifies these issues.

To more rigorously assess video synthesis quality, we additionally incorporate VideoMAE V2~\cite{videomae2paper} to measure the Fréchet Inception Distance for videos (video FID) or content-debiased FVD, extracting embeddings from 8 consecutive frames. This ensures a more robust evaluation of temporal coherence and perceptual fidelity than frame-by-frame metrics alone.

In this work, we address the decoder bottleneck by designing lightweight decoders that can seamlessly replace the original decoder in latent diffusion frameworks. Our approach aims to reduce both inference time and memory usage while maintaining acceptable reconstruction quality. This effort is aligned with a wider movement toward scalable and efficient generative models, as exemplified by VideoMAE V2~\cite{videomae2paper}, which employs dual masking strategies to improve efficiency in video pretraining.

We train our decoders on large-scale data subsets (e.g., LAION~\cite{laionface, laionaesth}) and evaluate them on COCO2017~\cite{coco2017} for images and UCF101~\cite{ucf101} for video tasks. Using smaller architectures (25--30\,MB in FP16) derived from diffusers, Vision Transformers~\cite{vit, efficientvit} and Taming Transformers~\cite{tamingtransformers}, we show that we can often preserve sufficient fidelity (as measured by SSIM, PSNR, and FID) while achieving up to 20$\times$ faster decoding at certain resolutions.

\section{Related Work}

\subsection{Diffusion Models}
Diffusion probabilistic models~\cite{sdxl, sddm} have become a standard approach in text-to-image and image-to-image generation, demonstrating training stability and the capability for photorealistic images. Stable Diffusion (SD) augments this approach with a powerful text encoder (e.g., CLIP) and a UNet denoising backbone that operates on a compressed latent space.

\subsection{Variational Autoencoders for Latent Diffusion}
Stable Diffusion employs a Variational Autoencoder (VAE~\cite{sdxl, sddm}) to facilitate the mapping of images to and from a latent space, which is a crucial component of its architecture. This innovative approach significantly reduces the complexity associated with the denoising process, allowing for more efficient computations and faster results. Using latent space, SD pipelines can effectively compress and reconstruct images, enhancing their ability to generate high-quality output. However, it is important to note that the default VAE decoders are often quite large ($\ge180$\,MB), which poses challenges during inference. The size of these decoders can lead to slower processing times and increased memory usage, which can be particularly problematic in resource-constrained environments or applications requiring real-time performance. As a result, optimizing the VAE architecture and exploring alternative decoder designs could be vital steps in improving the overall efficiency and usability of Stable Diffusion in various practical scenarios.

\subsection{Lightweight Vision Transformers}
Vision Transformers (ViTs) ~\cite{vit, efficientvit} have emerged as a powerful architecture in the realm of computer vision, demonstrating remarkable capabilities in both image recognition and generation tasks. Their unique attention-based mechanism allows them to capture long-range dependencies and intricate patterns within images, making them particularly effective for complex visual understanding. However, despite their impressive performance, ViTs can consume resources intensively, often requiring significant memory and computational power, which can limit their applicability in real-world scenarios, especially in environments with limited resources.

To address these challenges, researchers have been actively exploring various optimization strategies aimed at enhancing the memory and compute efficiency of Vision Transformers. Recent advances have led to the development of innovative variants such as EfficientViT, which significantly reduce the number of parameters while maintaining high levels of performance ~\cite{efficientvit}.

\subsection{Diffusion in Video}
Video diffusion models ~\cite{sddm, videomae2paper} represent a significant advance in the application of latent diffusion techniques to spatio-temporal data, effectively enabling the generation and manipulation of video content. These models leverage the principles of diffusion processes to iteratively refine video frames, allowing for the creation of visual sequences that can capture complex motion and temporal dynamics. However, one of the primary challenges associated with these models is the computational overhead incurred during the decoding process. Specifically, decoding each frame in a video sequence can be relatively expensive in terms of resources, particularly when dealing with lengthy content that spans several seconds or even minutes. This computational burden can lead to increased latency and reduced efficiency, which are critical factors to consider in applications requiring real-time processing or batch generation of videos.

To address this issue, the development of a lightweight video decoder becomes paramount. A streamlined decoder can significantly reduce the amount of computational resources required, enabling faster processing times without compromising the quality of the generated output. By optimizing the architecture and algorithms used in the decoding process, researchers can create solutions that are not only faster, but also more scalable, allowing for the handling of larger video datasets or more complex generation tasks. This is particularly important in contexts such as live streaming, video editing, and automated content creation, where responsiveness and efficiency are key. Furthermore, a lightweight decoder can facilitate the deployment of video diffusion models on edge devices, expanding their accessibility and applicability across various platforms and use cases. As the field of video generation continues to evolve, focusing on optimizing decoding processes will be essential for maximizing the potential of video diffusion models and ensuring their viability in practical applications.

\subsection{Scaling Video Masked Autoencoders with VideoMAE V2}
VideoMAE V2~\cite{videomae2paper} highlights the importance of scaling video masked autoencoders via a dual masking strategy (masking both the encoder and decoder). This progressive training approach demonstrates improved efficiency in video pre-training and opens new avenues for efficient generative architectures. Although VideoMAE V2 mainly focuses on the entire MAE framework rather than specifically on the decoder, its insights on efficiency and scaling complement our goal of building lightweight decoders for diffusion pipelines.

\subsection{Other Related Work}
In addition to the above, there are various lines of research exploring strategies to reduce the computational overhead of generative models. These include novel masking approaches (inspired by VideoMAE V2), progressive training on large-scale datasets, and advanced architectures for video tasks that better handle temporal consistency. Our work intersects with these efforts, offering a decoder-centric perspective.

\section{Method}

\subsection{Model Overview}
\begin{figure*}[t!]
    \centering
    \begin{subfigure}[t]{0.5\textwidth}
        \centering
        \includegraphics[height=2.3in]{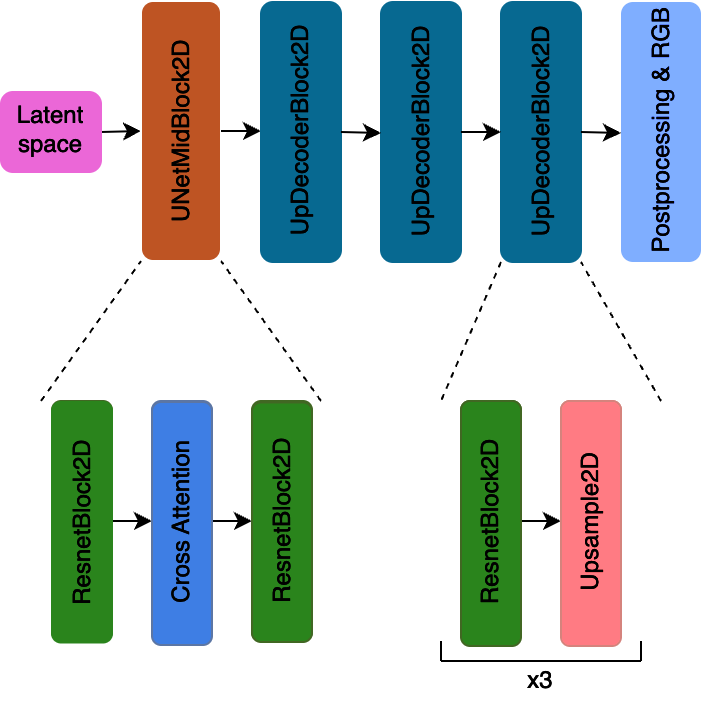}
        \caption{VAE decoder}
    \end{subfigure}%
    ~ 
    \begin{subfigure}[t]{0.5\textwidth}
        \centering
        \includegraphics[height=2.0in]{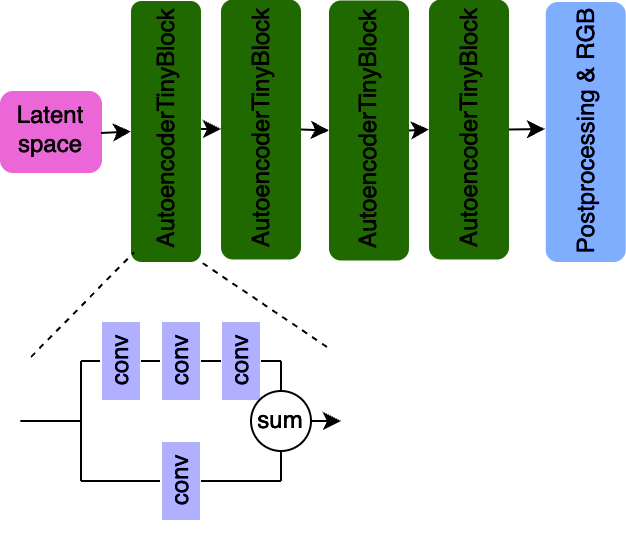}
        \caption{Tiny Decoder (from TAE)}
    \end{subfigure}
    
    \caption{Image decoder architectures from diffusers library}
    \label{fig:vae_image}
\end{figure*}

We modify only the decoder component of a standard latent diffusion model, keeping the UNet denoising processes intact. We experiment with two main lightweight decoder designs:
\begin{itemize}
    \item \textbf{TAE-192}: A small autoencoder architecture using diffusers and Taming Transformers~\cite{tamingtransformers} Python libraries.
    \item \textbf{EfficientViT}: A compressed Vision Transformer decoder derived from~\cite{efficientvit}.
\end{itemize}
Both are trained to reconstruct images (and video frames) from the corresponding latent codes generated by the original VAE encoder and presented in Figure \ref{fig:vae_image}.

\begin{figure*}[t!]
    \centering
    \begin{subfigure}[t]{0.5\textwidth}
        \centering
        \includegraphics[height=2.3in]{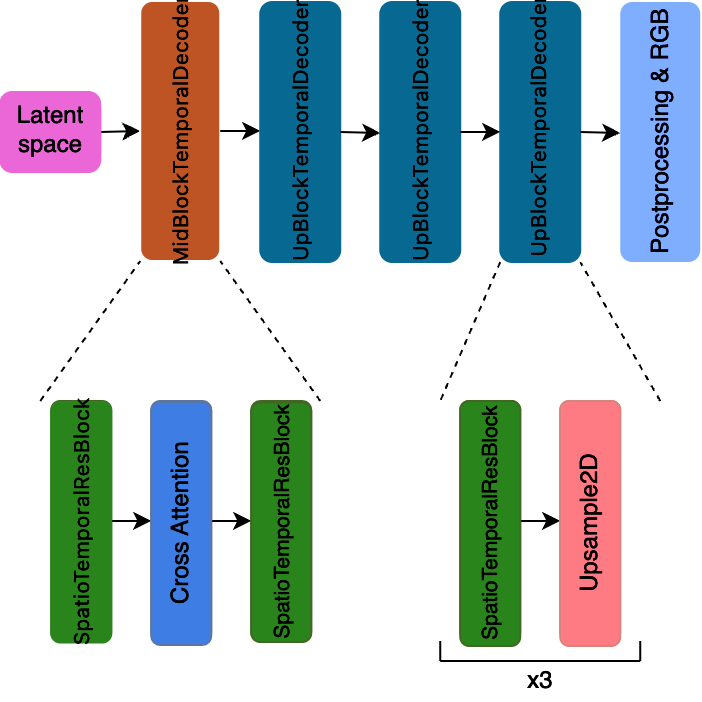}
        \caption{VAE video decoder}
    \end{subfigure}%
    ~ 
    \begin{subfigure}[t]{0.5\textwidth}
        \centering
        \caption{Tiny Temporal Decoder}
        \vspace*{-5cm}
        \includegraphics[height=1.2in]{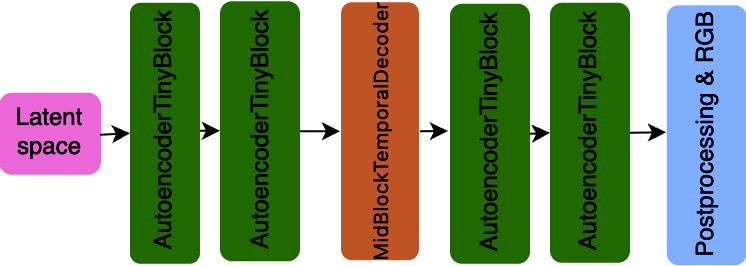}
    \end{subfigure}
    
    \caption{Video decoder architectures}
    \label{fig:vae_video}
\end{figure*}

For video models training, we additionally develop a ``TAE-192 Temporal'' variant, where we embed a mid spatio-temporal block inspired by the diffusers library, injecting temporal attention layers among the spatial ones. The corresponding video decoder architecture schemes are provided in Figure \ref{fig:vae_video}.

\subsection{Training Datasets}
Our training and evaluation pipeline uses the following datasets:
\begin{itemize}
    \item \textbf{LAION-Face}~\cite{laionface}: $\sim$20K images of diverse human faces to stress-test fidelity in facial reconstruction.
    \item \textbf{LAION-Aesthetics}~\cite{laionaesth}: $\sim140$K images selected by aesthetic scores (3--6 and 8). We sample 80K from the 3--6 range and 60K from the score of 8 subset.
    \item \textbf{COCO2017}~\cite{coco2017}: Used primarily for validation (val split, containing 5000 images).
    \item \textbf{UCF101}~\cite{ucf101}: For video experiments, we select $\sim$40--45 categories (e.g., BenchPress, Punch, SkyDiving, CricketShot) to validate temporal reconstruction.
\end{itemize}
In addition, we incorporate portions of WebVid for training the TAE-192 Temporal model on diverse short video clips, aiming to improve generalization for subsequent UCF-101 experiments.

\subsection{Decoder Training Strategy}
We use the original VAE encoder from Stable Diffusion~\cite{sdxl} to produce latent representations, and train our lightweight decoder to invert these latents back to RGB frames. We employ a combined reconstruction loss:
\begin{equation}
L \;=\; \alpha \, L_{\mathrm{MSE}} \;+\; \beta \, L_{\mathrm{LPIPS}} \;+\; \gamma \, L_{\mathrm{GAN}} \; \Theta(t-t_0),
\end{equation}
where $L_{\mathrm{MSE}}$ enforces pixel-level fidelity, $L_{\mathrm{LPIPS}}$~\cite{lpips} encourages perceptual similarity, and $L_{\mathrm{GAN}}$~\cite{gan} provides adversarial supervision and $\Theta(t-t_0)$ is the Heaviside step function, which we use to enable adversarial loss $L_{GAN}$ from the training step $t_0$. In some experiments, we assign higher weights to facial regions to improve their quality, leveraging LAION-Face.
We set $\alpha=1.0$, $\beta=0.4$, $\gamma=0.8$ and $t_0=10000$. 

We train models on 8 A100 GPUs (40\,GB) for 14 epochs, with progressive data augmentation for images and similar but slightly modified settings for video frames. For video, a temporal alignment error term ($L_{\mathrm{TAE}}$) is added to maintain consistency between consecutive frames.

\subsection{Potential Exploration of Dual Masking}
Inspired by the Dual Masking Strategy of VideoMAE V2, future work could investigate the application of partial masking to the input of the decoder during inference. By reducing the volume of features to be decoded, we may further lower computational overhead. Although this paper focuses on architectural modifications, the integration of masking could be a promising follow-up.

\subsection{Video Quality Metrics (Using VideoMAE V2 Embeddings)}
To better evaluate the perceptual quality of generated videos, we augment our pipeline with the VideoMAE V2~\cite{videomae2paper} backbone to extract embeddings for computing advanced metrics such as video FID or content-debiased FVD. Specifically, we sample 8 consecutive frames, then feed them into VideoMAE V2, and compute Fréchet-like distances in the resulting embedding space. This approach captures temporal coherence and content-level fidelity more effectively than frame-by-frame metrics alone.

In our approach, we meticulously sample eight consecutive frames from the generated video, ensuring that we capture a representative slice of the temporal dynamics present in the content. These sampled frames are then fed into the VideoMAE V2 model, which processes them to produce high-dimensional embeddings that encapsulate both spatial and temporal features. By computing Fréchet Inception Distance within this resulting embedding space, we can gain deeper insights into the relationships between generated videos and reference videos, allowing us to quantify how closely they align in terms of perceptual quality.

\section{Experiments and Results}

\subsection{Evaluation Metrics}
 The introduction of robust evaluation metrics is crucial for models performance assessing. This paper employs five key metrics:
\begin{itemize}
    \item \textbf{SSIM} ($\uparrow$): Structural Similarity Index Measure evaluates the perceptual similarity between two images by considering luminance, contrast, and structure, providing insights into how closely the generated images resemble the ground truth. 
    \item \textbf{PSNR} ($\uparrow$): Peak Signal-to-Noise Ratio quantifies the ratio between the maximum possible power of a signal and the power of distorting noise, serving as a straightforward measure of image quality.
    \item \textbf{FID} ($\downarrow$): Fréchet Inception Distance measures the distance between feature distributions of real and generated images in a high-dimensional feature space.
    \item \textbf{Runtime} ($\Delta t$): Average decoding time per image/bunch of frames (seconds).
    \item \textbf{Memory Footprint} (MB): Approximate model size on disk.
\end{itemize}

\subsection{Image Results on COCO2017}
\begin{figure}[h]
  \centering
  \includegraphics[width=0.9\linewidth]{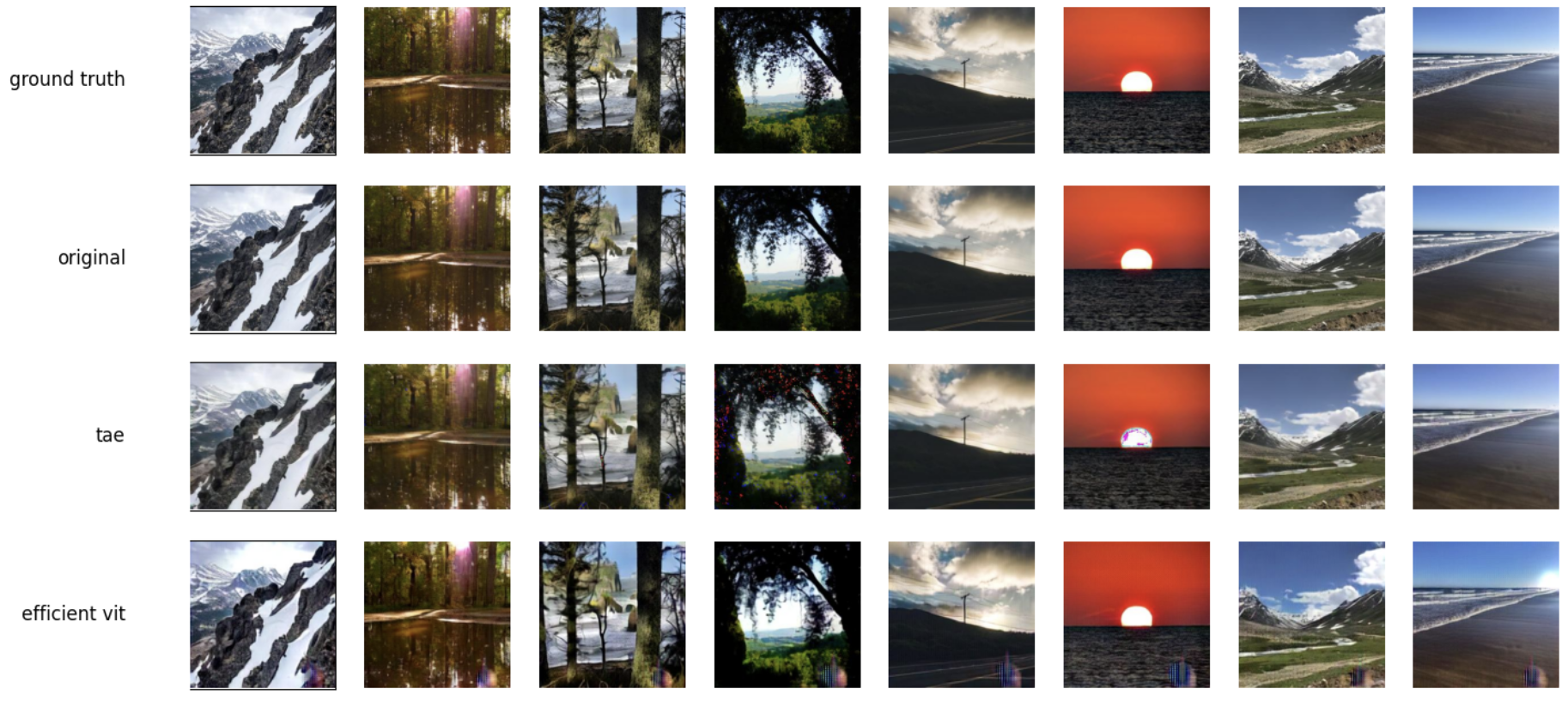}
  \caption{Example reconstructions on COCO2017 validation split, resolution 256x256. Top row: ground truth; second row: SDXL VAE; third row: TAE-192; fourth row: EfficientViT.}
  \label{fig:coco-image-compare}
\end{figure}

We evaluate three image decoders (baseline SDXL VAE, TAE-192, EfficientViT) at $256\times256$ and $1024\times1024$ resolutions. Representative results (mean $\pm$ std, where available) are shown below, reconstructions of the resolution 256x256 are shown in Figure \ref{fig:coco-image-compare}.

\paragraph{$256\times256$ Resolution.}

\begin{table}[h]
\centering
\caption{$256\times256$ decoding performance on COCO2017}
\label{tab:256-coco}
\begin{tabular}{@{}lcccc@{}}
\toprule
\textbf{Decoder} & \textbf{SSIM} & \textbf{PSNR} & \textbf{FID} & \textbf{$\Delta t$ (sec)} \\
\midrule
SDXL VAE~\cite{sdxl,sddm}     & 0.7656 $\pm$ 0.0023 & 25.72 $\pm$ 0.09 & 2.2295  & 0.0100 \\
TAE-192                      & 0.7034 $\pm$ 0.0029 & 21.8643 $\pm$ 0.03 & 21.4765 & 0.0047 \\
EffViT                       & 0.5483 $\pm$ 0.0015 & 20.49 $\pm$ 0.03 & 25.0572 & 0.0069 \\
\bottomrule
\end{tabular}
\end{table}

While TAE-192 and EfficientViT yield lower SSIM/PSNR than the baseline, they reduce decoding time by nearly half (\(0.0100 \rightarrow \sim 0.0047{-}0.0069\)).

\paragraph{$1024\times1024$ Resolution.}

\begin{table}[h]
\centering
\caption{$1024\times1024$ decoding performance on COCO2017}
\label{tab:1024-coco}
\begin{tabular}{@{}lcccc@{}}
\toprule
\textbf{Decoder} & \textbf{SSIM} & \textbf{PSNR} & \textbf{FID} & \textbf{$\Delta t$ (sec)} \\
\midrule
SDXL VAE~\cite{sdxl} & 0.7729 $\pm$ 0.0024 & 26.19 $\pm$ 0.11 & 1.1004  & 0.0094 \\
TAE-192             & 0.7497 $\pm$ 0.0025 & 25.47 $\pm$ 0.09 & 2.4810  & 0.00463 \\
EffViT              & 0.6427 $\pm$ 0.0024 & 21.95 $\pm$ 0.05 & 7.6543  & 0.00701 \\
\bottomrule
\end{tabular}
\end{table}

Both custom decoders run about 1.5--2$\times$ faster at $1024\times1024$. TAE-192 preserves perceptual quality in a way better than EffViT (closer to the baseline in SSIM/PSNR), albeit with a higher FID than the original VAE.

\subsection{Video Results on UCF-101 Subset}
We compare the standard SVD VAE from~\cite{sdxl,videomae2paper} (referred to here as ``SVD VAE'') with our approaches on around 40 classes from UCF-101:

\begin{table}[h]
\centering
\caption{Video decoding results on UCF-101 (subset).}
\label{tab:ucf-video}
\begin{tabular}{@{}lcc@{}}
\toprule
\textbf{Decoder} & \textbf{$\Delta t$ (sec/8 frames)} & \textbf{Video FID} \\
\midrule
SVD VAE~\cite{sdxl,videomae2paper} & 0.02169 & 8.2604 \\
TAE-192 (not temporal)                 & 0.00424 & 76.8414 \\
TAE-Temporal                       & 0.00771 & 19.2186 \\
\bottomrule
\end{tabular}
\end{table}

A temporal version of TAE-192 drastically reduces decoding time (0.00424\,sec/frame) but yields a high FID (76.84). Adding a Mid SpatioTemporal block to the TAE-192 architecture balances speed and quality (FID is 19.22 at 0.00771\,sec/frame). This underscores the trade-off between speed and perceptual quality, especially in video.

We also evaluated these models using VideoMAE V2 embeddings for an additional Video FID measure 
, confirming that temporal consistency degrades significantly without temporal layers and alignment terms.

\begin{figure}[h]
  \centering
  \includegraphics[width=0.95\linewidth]{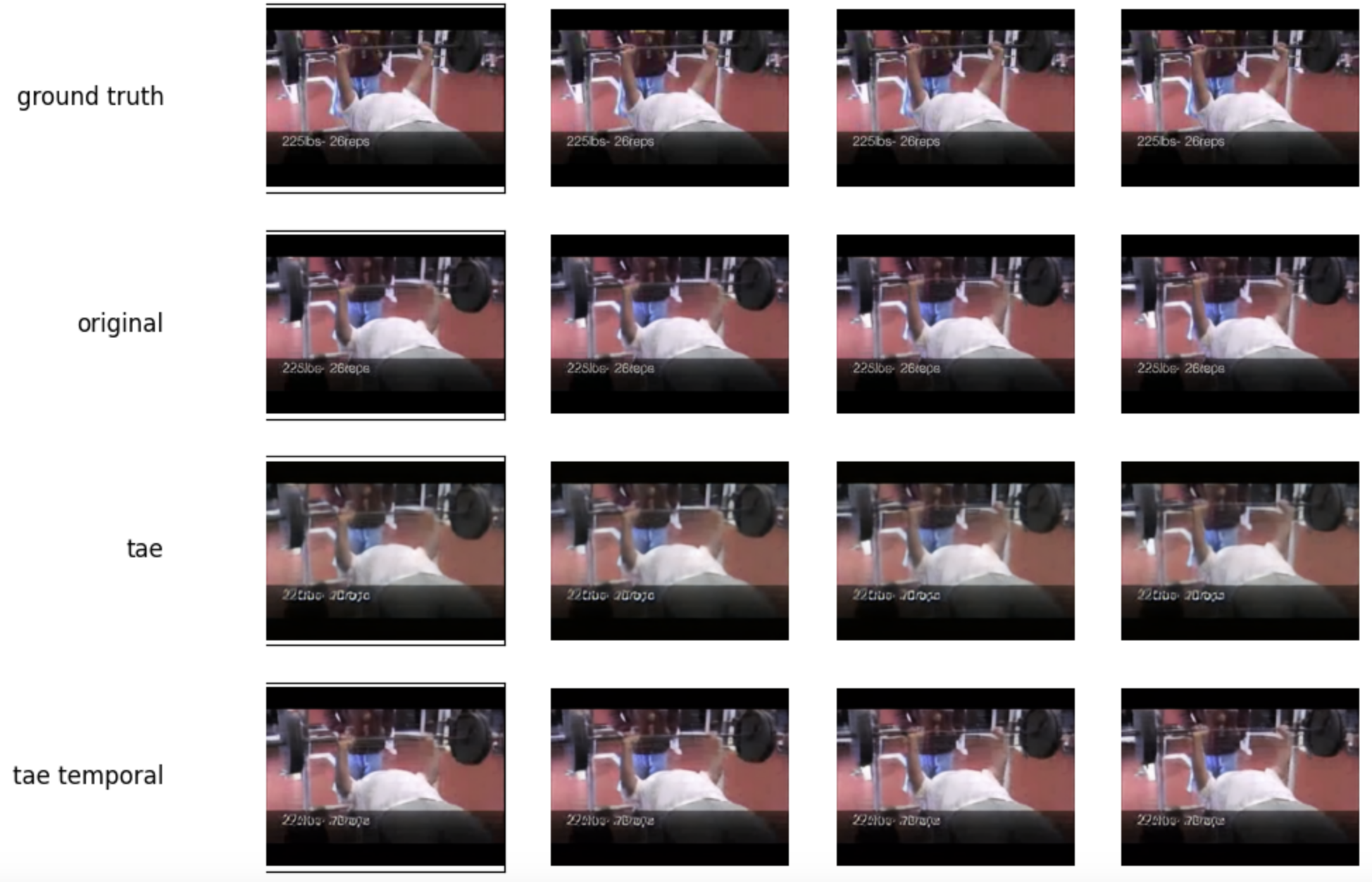}
  \caption{Frames from UCF101: top row, ground-truth; second row, SVD VAE; third row, TAE-192 (temporal); bottom row, TAE-Temporal. More examples shown in Appendix.}
  \label{fig:ucf-video-compare}
\end{figure}

\subsection{Ablation on Decoder Masking (Inspired by VideoMAE V2)}
Though not implemented in our current models, we draw on VideoMAE V2's masking ablation for potential future work. This could include:
\begin{itemize}
    \item Varying masking ratios during inference to measure the speed--quality trade-off.
    \item Exploring random vs.\ structured masking strategies.
\end{itemize}

\subsection{Potential Comparison with VideoMAE V2}
A direct comparison on UCF-101 or another benchmark could quantify differences between our lightweight decoder approach and VideoMAE V2's dual masking strategy. Metrics of interest would include speed, memory usage, and generation quality under similar experimental conditions.

\section{Discussion}
Our experiments show that replacing the default VAE decoder with smaller, custom architectures can substantially accelerate the final step of latent diffusion pipelines for both images and video. Although these decoders yield lower SSIM/PSNR scores compared to the baseline, the speed and memory benefits can be critical for large-scale or near-real-time tasks.

For video, a pure training strategy without temporal constraints can degrade perceptual consistency across frames. Incorporating temporal alignment losses or adversarial terms partially mitigates this. Future work will extend these approaches, investigating decoders that natively handle spatiotemporal dynamics with balanced speed, memory footprint, and visual fidelity.

\subsection{Generalizability and Future Work}
VideoMAE V2 demonstrates the potential of masked autoencoders across a range of video tasks (e.g., spatial/temporal action detection). Examining how our lightweight decoders generalize to these tasks could further validate their utility. Additionally, integrating dual masking or adopting more elaborate progressive training paradigms may amplify the efficiency benefits.

\subsection{Scaling Challenges}
VideoMAE V2's successes with billion-parameter models highlight the continuing challenge of scaling video architectures. Our lightweight decoders address part of this challenge by easing the decoder's computational cost. In principle, this could free up resources for larger UNet or encoder modules, supporting more complex tasks or longer video sequences.

\subsection{Data Efficiency}
VideoMAE V2 emphasizes data efficiency through masking and progressive training. In future work, we plan to explore how lightweight decoders might further reduce training data requirements or accelerate the training schedule for video generation.

\section{Conclusion}
We present a systematic exploration of lightweight decoders for latent diffusion-based image and video generation. Training TAE and EfficientViT decoders on LAION-Face, LAION-Aesthetics, and subsets of UCF101 yields up to 20$\times$ faster decoding and 10--15\% total speed-ups in certain pipelines, with moderately reduced memory usage. While perceptual quality can drop relative to standard VAEs, such trade-offs are often worthwhile in large-scale or real-time settings.

By building on insights from VideoMAE V2 regarding efficiency and scalability, we see multiple directions for future research, including dual masking strategies, progressive training approaches, and integration with billion-parameter video models. We hope that publicly releasing our decoder checkpoints and code will inspire further optimization and innovation in efficient latent diffusion architectures.

\subsection*{Acknowledgments}
We thank the maintainers of open-source libraries and datasets---Hugging Face Diffusers, Stability AI, LAION, COCO, and UCF101---for enabling rapid experimentation. We also acknowledge the VideoMAE V2 authors for providing valuable insights into efficiency and scalability, which have informed our work.

\appendix
\section*{Appendix / Code and Reproducibility}
We will release code and pretrained checkpoints for our lightweight decoders, along with integration scripts for existing Stable Diffusion pipelines, in a public GitHub repository upon publication. We note that certain pretrained models may be subject to licensing constraints. Where necessary, we will provide partial weights or configuration files to facilitate further research while respecting any dataset- or model-specific usage restrictions.

\begin{figure}[h]
  \centering
  \includegraphics[width=0.95\linewidth]{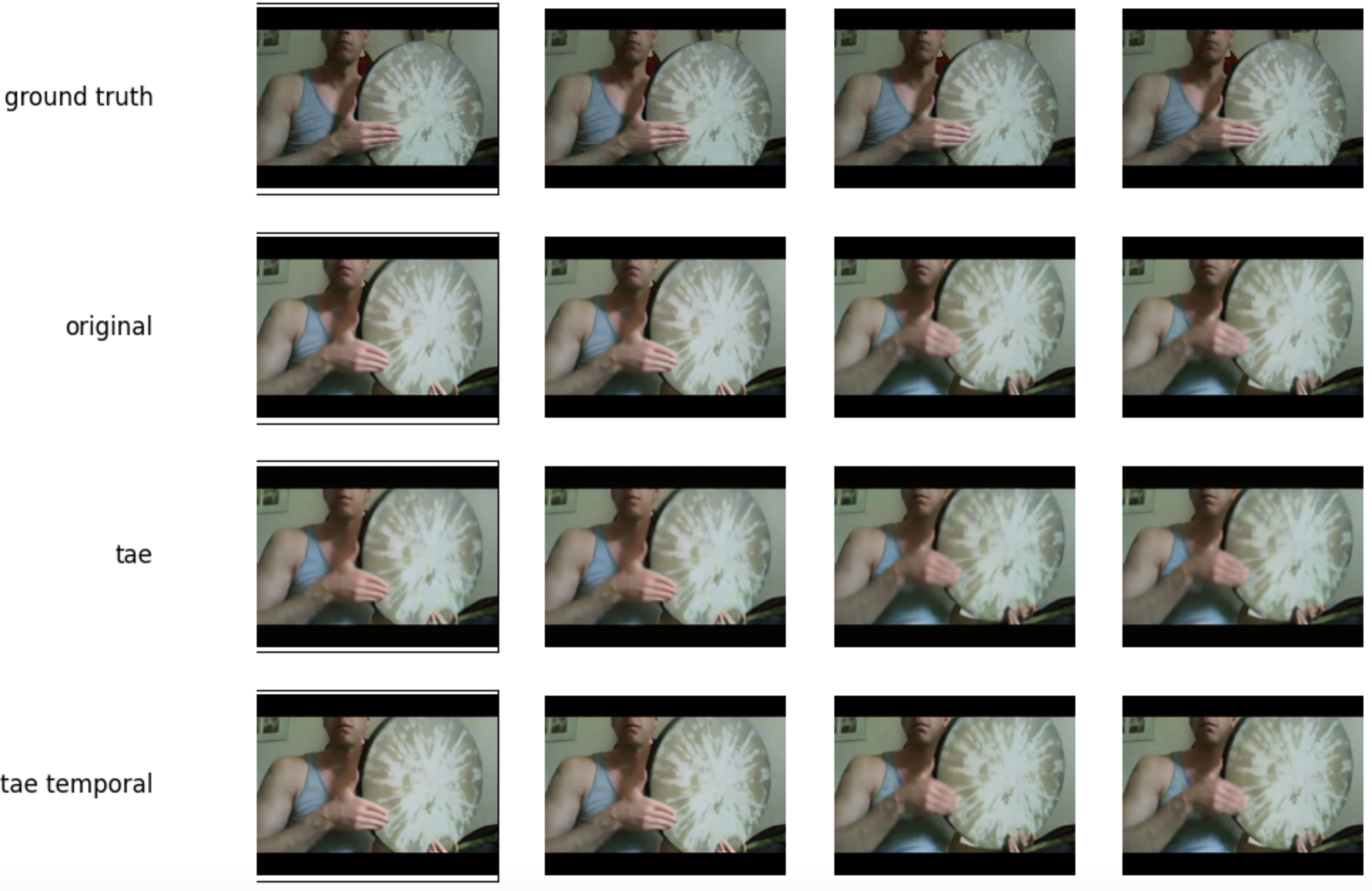}
  \caption{Additional video-decoder examples on UCF101, part 1.}
\end{figure}

\begin{figure}[h]
  \centering
  \includegraphics[width=0.95\linewidth]{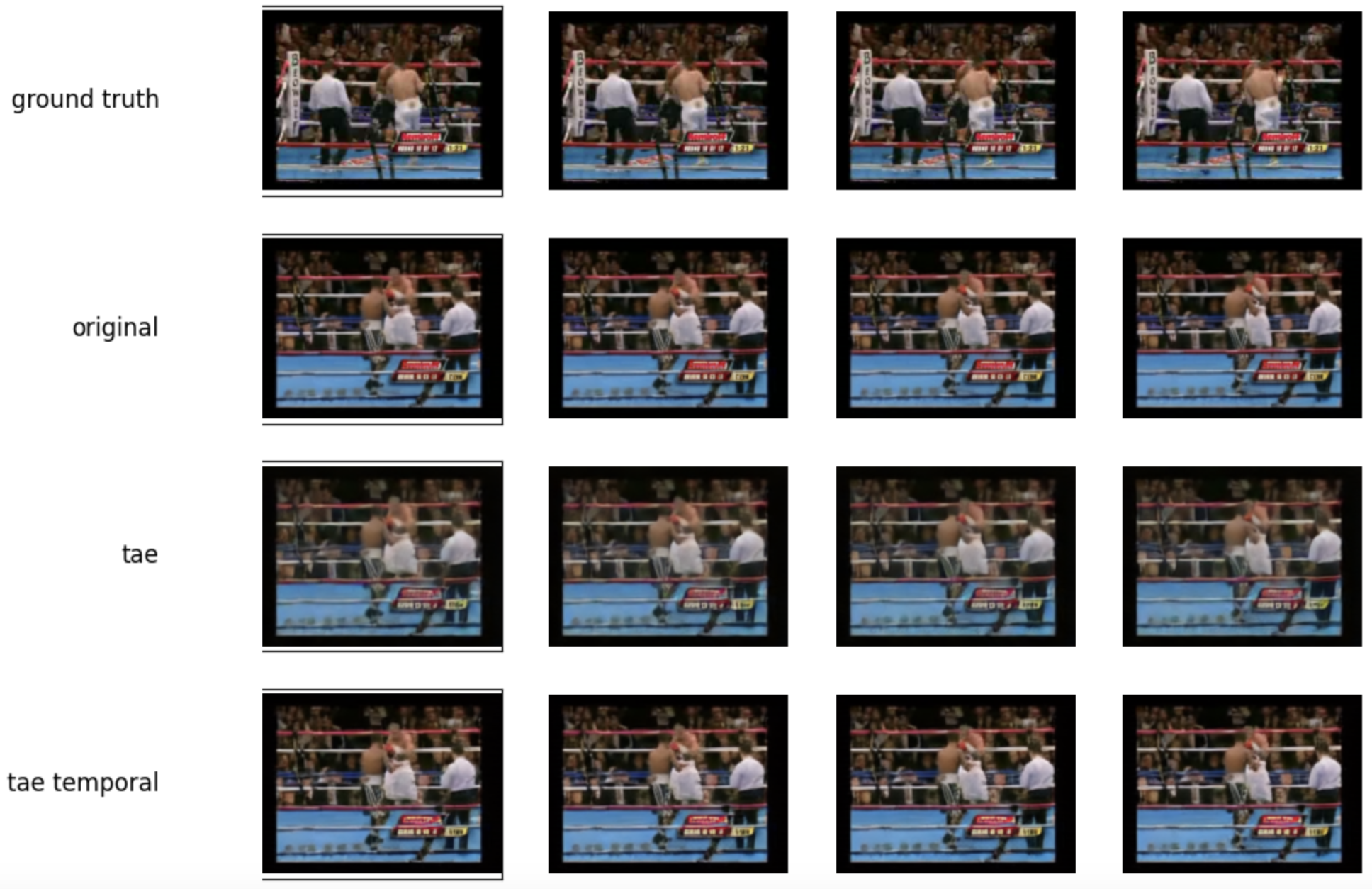}
  \caption{Additional video-decoder examples on UCF101, part 2.}
\end{figure}

\begin{figure}[h]
  \centering
  \includegraphics[width=0.95\linewidth]{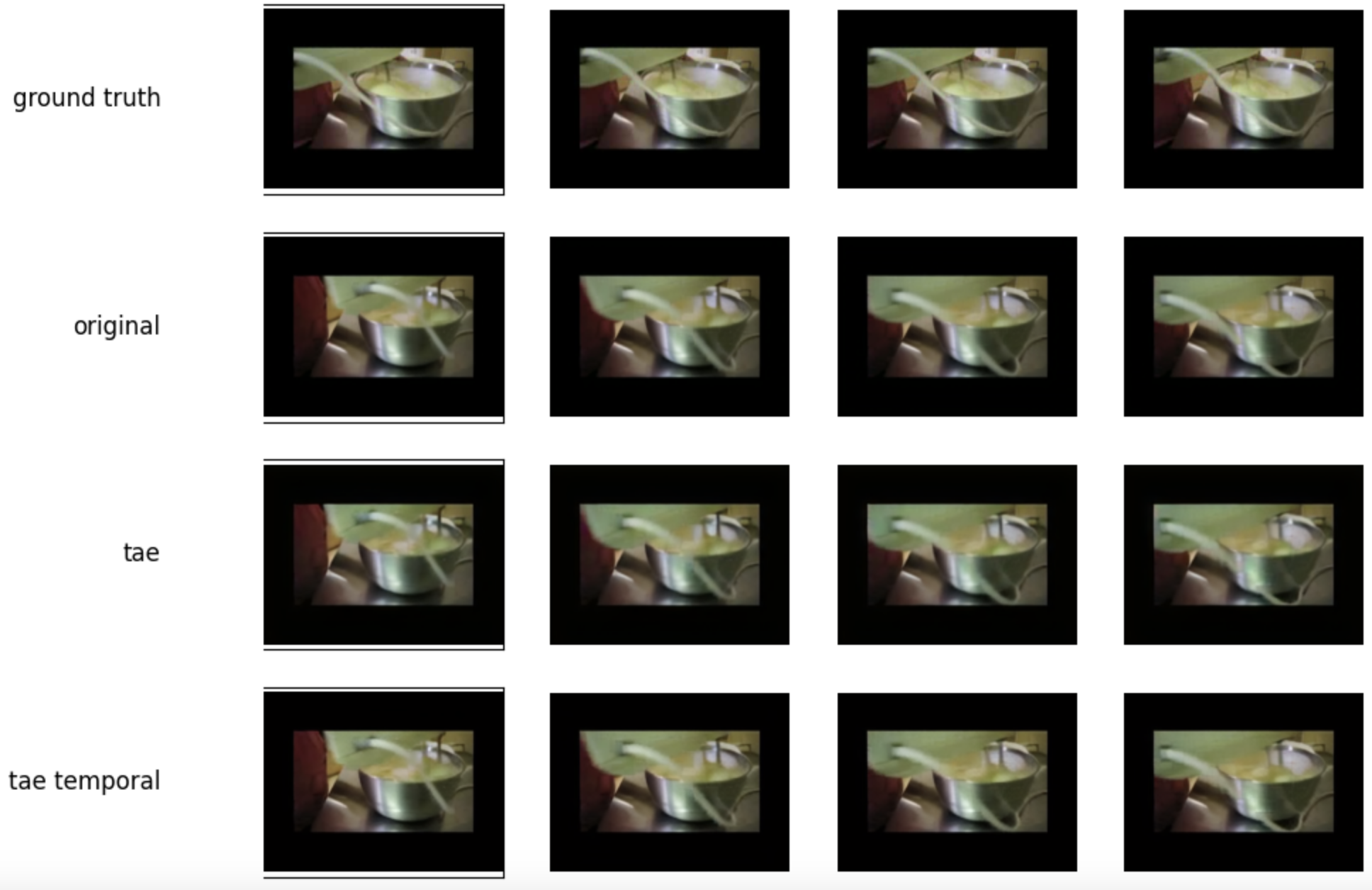}
  \caption{Additional video-decoder examples on UCF101, part 3.}
\end{figure}

\begin{figure}[h]
  \centering
  \includegraphics[width=0.95\linewidth]{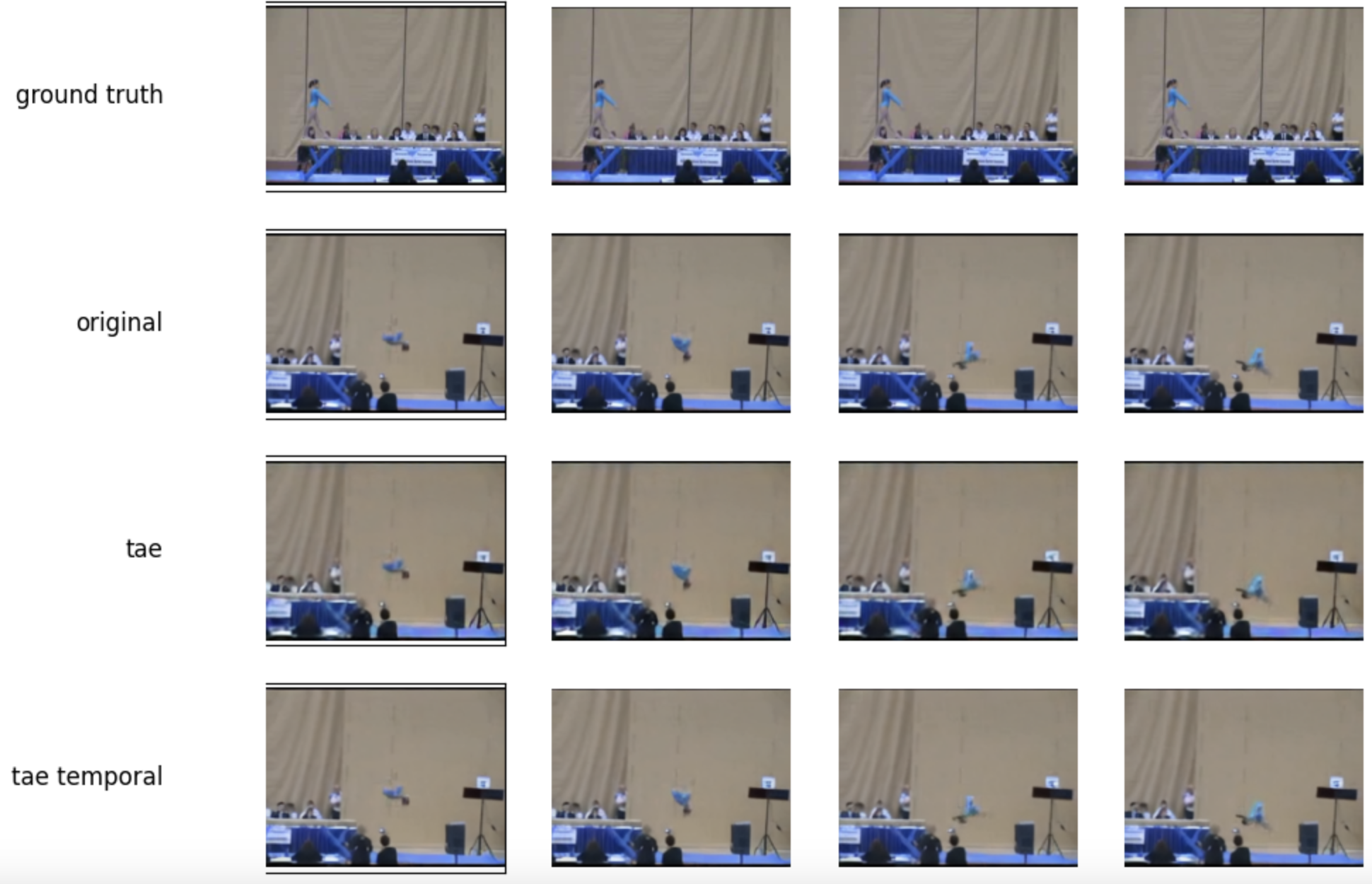}
  \caption{Additional video-decoder examples on UCF101, part 4.}
\end{figure}

\end{document}